\documentclass{article} 
\usepackage{iclr2025_conference,times}

\usepackage{amsmath,amsfonts,bm}

\def\eqref#1{equation~\ref{#1}}

\def\1{\bm{1}}

\DeclareMathAlphabet{\mathsfit}{\encodingdefault}{\sfdefault}{m}{sl}
\SetMathAlphabet{\mathsfit}{bold}{\encodingdefault}{\sfdefault}{bx}{n}

\usepackage{hyperref}
\usepackage{url}
\usepackage{graphicx} 
\usepackage{caption} 
\usepackage{cleveref}
\usepackage{booktabs}
\usepackage{amssymb}
\usepackage{multirow}
\usepackage{wrapfig}

\definecolor{LightBlue}{rgb}{0.9,0.94,1}

\title{FastFit: Accelerating Multi-Reference Virtual Try-On via Cacheable Diffusion Models}

\author{
Zheng Chong\textsuperscript{1,2,3}, 
Yanwei Lei\textsuperscript{1}, 
Shiyue Zhang\textsuperscript{1}, 
Zhuandi He\textsuperscript{1}, 
Zhen Wang\textsuperscript{1}, 
Xujie Zhang\textsuperscript{1}, \\
\hspace{0.2mm}
\textbf{Xiao Dong\textsuperscript{1}}, 
\textbf{Yiling Wu}\textsuperscript{\textbf{3}},
\textbf{Dongmei Jiang}\textsuperscript{\textbf{3}} \& 
\textbf{Xiaodan Liang}\textsuperscript{\textbf{1,3}}\thanks{Corresponding author. Project page: \href{https://github.com/Zheng-Chong/FastFit}{https://github.com/Zheng-Chong/FastFit}.}
\\
\textsuperscript{1}Sun Yat-sen University  \hspace{2mm}
\textsuperscript{2}LavieAI \hspace{2mm}
\textsuperscript{3}Pengcheng Laboratory \\
\small\texttt{\{chongzheng98,dx.icandoti,xdliang328\}@gmail.com, } \\ 
\small\texttt{\{leiyw5,zhangshy223,zhuandihe86,wangzh669,zhangxj59\}@mail2.sysu.edu.cn,} \\ 
\small\texttt{\{wuyl02,jiangdm\}@pcl.ac.cn}
\vspace{2mm}\\
}

\iclrfinalcopy 
\begin{document}

\maketitle
\vspace{-6mm}

\captionsetup{type=figure}

\hspace{2mm}
\includegraphics[width=0.95\textwidth]{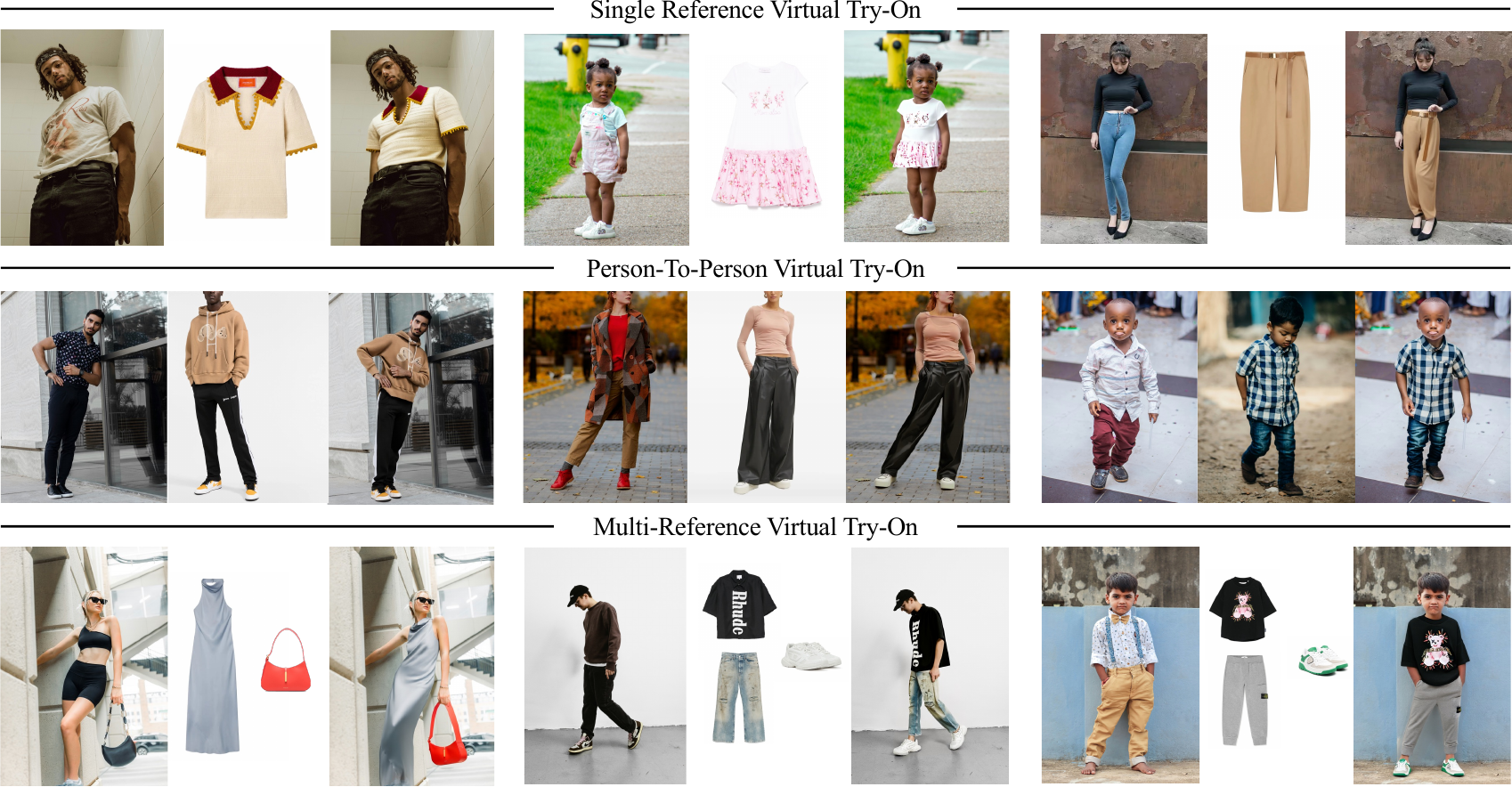}

\caption{FastFit provides a unified and accelerated solution for diverse virtual try-on tasks, including single-reference, person-to-person, and our primary focus, multi-reference composition. By decoupling the reference images from the denoising process, our cacheable diffusion architecture delivers high-fidelity virtual try-on across multiple challenging scenarios at a much faster speed.}
\label{fig:teaser}
\vspace{2mm}

\begin{abstract}
Despite its great potential, virtual try-on technology is hindered from real-world application by two major challenges: the inability of current methods to support multi-reference outfit compositions (including garments and accessories), and their significant inefficiency caused by the redundant re-computation of reference features in each denoising step. To address these challenges, we propose FastFit, a high-speed multi-reference virtual try-on framework based on a novel cacheable diffusion architecture.
By employing a Semi-Attention mechanism and substituting traditional timestep embeddings with class embeddings for reference items, our model fully decouples reference feature encoding from the denoising process with negligible parameter overhead. This allows reference features to be computed only once and losslessly reused across all steps, fundamentally breaking the efficiency bottleneck and achieving an average 3.5$\times$ speedup over comparable methods.
Furthermore, to facilitate research on complex, multi-reference virtual try-on, we introduce DressCode-MR, a new large-scale dataset. It comprises 28,179 sets of high-quality, paired images covering five key categories (tops, bottoms, dresses, shoes, and bags), constructed through a pipeline of expert models and human feedback refinement. 
Extensive experiments on the VITON-HD, DressCode, and our DressCode-MR datasets show that FastFit surpasses state-of-the-art methods on key fidelity metrics while offering its significant advantage in inference efficiency.
\end{abstract}

\vspace{3mm}

\section{Introduction}

\begin{wrapfigure}{tr}{0.45\textwidth}
    \vspace{-13pt} 
    \centering
    \includegraphics[width=\linewidth]{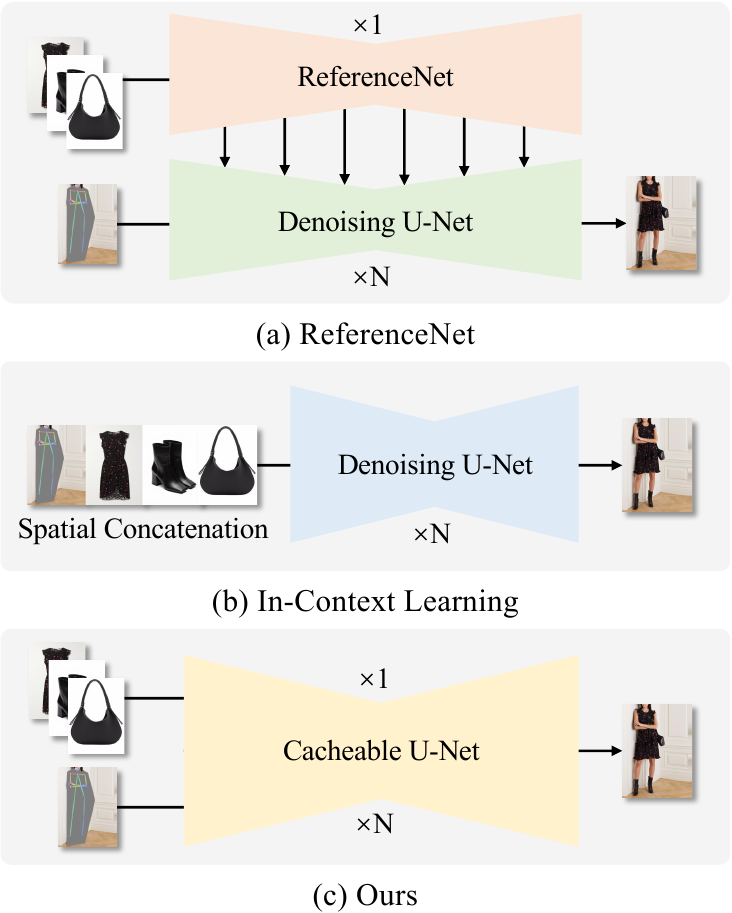}
    \caption{Architectural comparison of multi-reference try-on methods. Our cacheable U-Net (c) avoids the parameter overhead of ReferenceNet (a) and the computational redundancy of In-Context Learning (b).}
    \label{fig:structure_compare}
    \vspace{-11pt} 
\end{wrapfigure}

Generative AI-based virtual try-on has recently made remarkable progress. An ideal virtual try-on system—one that could revolutionize online retail and power applications like intelligent outfit visualization—would allow users to seamlessly mix and match various garments and accessories, rapidly generating photorealistic results to enable an interactive experience.
However, two major challenges hinder current methods from achieving this vision. 
Firstly, most existing methods \citep{xie2022pastaganversatileframeworkhighresolution, wang2018cpvton, xu2024ootdiffusion, choi2024idmvton, chong2024catvtonconcatenationneedvirtual, jiang2024fitditadvancingauthenticgarment} are designed for a single reference garment (e.g., a top or a dress), requiring a complete multi-item outfit to be rendered through iterative passes, leading to both inflated computation time and the risk of accumulated synthesis errors. Furthermore, the general lack of support for essential accessories like shoes and bags prevents the generation of truly holistic and realistic outfits.
Secondly, the computational inefficiency of current methods stems from two competing yet flawed strategies, as illustrated in \Cref{fig:structure_compare}. 
On one hand, ReferenceNet-based methods~\citep{huang2024parts2whole, choi2024idmvton, xu2024ootdiffusion, zhang2024mmtryonmultimodalmultireferencecontrol, zhou2024learning, jiang2024fitditadvancingauthenticgarment} employ a separate network to encode references (\Cref{fig:structure_compare} (a)), which avoids this redundancy but at the cost of substantial parameter overhead, increasing both training and inference costs. 
On the other hand, in-context learning-based methods~\citep{guo2025any2anytryonleveragingadaptiveposition, chong2024catvtonconcatenationneedvirtual, huang2024incontextlora} repeatedly process the concatenated reference and person features at each of the $N$ denoising steps (\Cref{fig:structure_compare} (b)), causing significant computational redundancy.

\vspace{1.5mm}
To overcome these limitations, we introduce FastFit, a high-speed framework that enables coherent multi-reference virtual try-on through a novel cacheable diffusion architecture. 
Our proposed Cacheable UNet decouples the reference feature encoding from the iterative denoising process, which is achieved by introducing a Reference Class Embedding and a Semi-Attention mechanism. 
This structure enables a Reference KV Cache during inference, which allows reference features to be computed only once and losslessly reused in all subsequent steps, fundamentally breaking the efficiency bottleneck and achieving an average 3.5$\times$ speedup over comparable methods with negligible parameter overhead.
Furthermore, observing the lack of datasets with complete outfit pairings, we construct DressCode-MR, a large-scale multi-reference try-on dataset based on \cite{morelli2022dresscode}. We developed a data-generation pipeline that trains expert models based on ~\cite{chong2024catvtonconcatenationneedvirtual} and \cite{flux2024} to recover canonical images of individual items, and utilizes human feedback to ensure high quality. This results in 28,179 multi-reference image sets spanning five key categories: tops, bottoms, dresses, shoes, and bags.

\vspace{1.5mm}
In summary, the contributions of this work include:
\begin{itemize} 
\item We propose FastFit, a novel framework for high-speed, multi-reference virtual try-on. It is the first to enable coherent  multi-reference virtual try-on across five key categories, including tops, bottoms, dresses, shoes, and bags, while achieving an average 3.5$\times$ speedup over comparable methods.
\item We design a novel Cacheable UNet structure featuring a Reference Class Embedding and a Semi-Attention mechanism. This design decouples reference feature encoding from the denoising process, enabling a lossless Reference KV Cache that breaks the core efficiency bottleneck of subject-driven generation architectures. 
\item We construct DressCode-MR, the first large-scale dataset specifically for multi-reference virtual try-on. It comprises 28,179 high-quality image sets, providing a solid foundation to foster future research in complex outfit generation. 
\item We conduct extensive experiments on VITON-HD, DressCode, and our DressCode-MR benchmarks, demonstrating that FastFit surpasses state-of-the-art methods in image fidelity while maintaining its significant efficiency advantage. 

\end{itemize}

\begin{figure*}
    \centering
    \includegraphics[width=\textwidth]{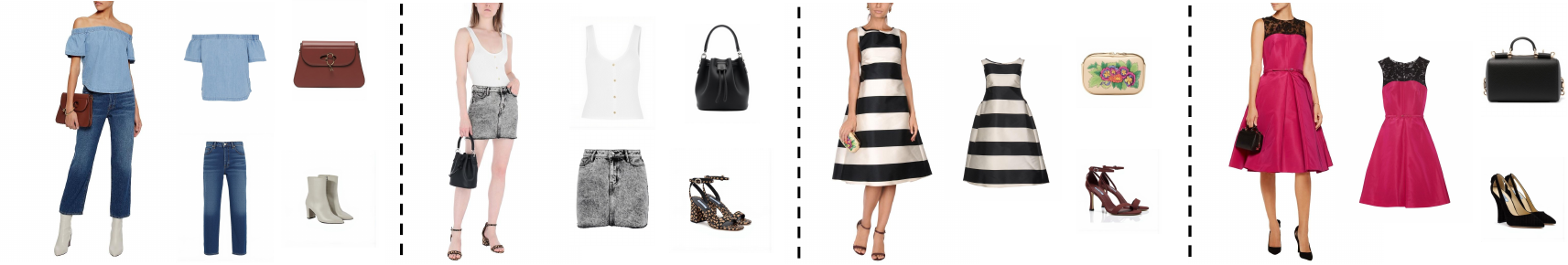}
    \vspace{-5mm}
    \caption{Illustrative examples from our proposed DressCode-MR dataset. Each sample provides a pairing of a full-body person image with a set of corresponding canonical images for each item. }
    \label{fig:dataset}
\end{figure*}

\section{Related Work}
\subsection{Subject-Driven Image Generation}

To enable finer-grained control in diffusion models for image generation, the research community has rapidly shifted towards subject-driven image generation. 
Early efforts primarily centered on single reference images, injecting specific subject identities or artistic styles by fine-tuning model weights \citep{ruiz2022dreambooth, yang2022paintbyexample, hu2021loralowrankadaptationlarge, huang2024incontextlora} or utilizing lightweight adapters \citep{ye2023ip-adapter, mou2023t2iadapter, chen2023anydoor}. However, the former approach requires training a separate model for each subject, limiting its practical flexibility, while the latter, despite being convenient, often faces challenges in maintaining high fidelity to the reference image. 
Another line of work based on in-context learning, such as IC-LoRA \citep{huang2024incontextlora} and OminiControl \citep{tan_ominicontrol2_2025, tan2025ominicontrol}, achieves superior detail preservation by concatenating the reference image with noise along the spatial dimension. The trade-off is that the reference must participate in every denoising step, significantly increasing inference time and computational cost. The limitations of these single-reference approaches become apparent when creative needs involve composing elements from multiple, diverse sources. Consequently, some works have begun to explore multi-reference generation; for instance, IC-Custom \citep{li2025iccustomdiverseimagecustomization} inputs multiple images as a single concatenated map for multi-concept composition, Face-diffuser investigates the complex multi-person synthesis task, and MultiRef \citep{chen2025multiref} provides the first systematic definition and benchmark for this task. Nevertheless, in the domain of virtual try-on, multi-reference generation remains an under-explored area. How to harmoniously compose visual information from multiple references while mitigating the heightened computational burden from increased inputs remains a significant and open challenge.

\subsection{Image-based Virtual Try-On}

Image-based virtual try-on aims to realistically synthesize a person wearing target garments. Classic paradigms centered on a warp-and-fuse method, which explicitly deforms the garment using either geometric transformations or learned appearance flows before the blending stage \citep{wang2018cpvton, han2017viton, choi2021vitonhd, han2019clothflow, ge2021pfafn, xie2021wasvtonwarpingarchitecturesearch, xie2023gpvton, gou2023dcivton, chong2022st-vton}; however, these approaches are frequently hampered by visual artifacts from inaccurate warping. Subsequently, the advent of diffusion models revolutionized the field by reframing the task as end-to-end conditional image generation, bypassing the error-prone warping step. The dominant strategy in these modern models involves injecting high-fidelity garment features into the denoising process via sophisticated conditioning mechanisms, such as parallel encoder branches (i.e., ReferenceNets) or ControlNet\citep{zhang2023adding}-like structures, a technique employed by a vast body of recent work \citep{zhu2023tryondiffusion, morelli2023ladi-vton, kim2023stableviton, xu2024ootdiffusion, wang2024stablegarment, choi2024idmvton, sun2024outfitanyoneultrahighqualityvirtual, zhou2024learning, zhang2024boowvtonboostinginthewildvirtual, kim2024promptdresserimprovingqualitycontrollability}. Recent innovations further push the boundaries by exploring alternative backbones like Diffusion Transformers \citep{peebles2022scalabledit} or introducing novel control modalities such as textual prompts and more generalized conditioning schemes \citep{ guo2025any2anytryonleveragingadaptiveposition, jiang2024fitditadvancingauthenticgarment}. Despite achieving unprecedented realism, their inference speed and general limitation to single garments have become key bottlenecks, hindering the technology's application in real-world scenarios that demand rapid feedback and multi-item outfit composition.

\section{Methods}

\begin{figure*}
  \centering
  \includegraphics[width=1.0\textwidth]{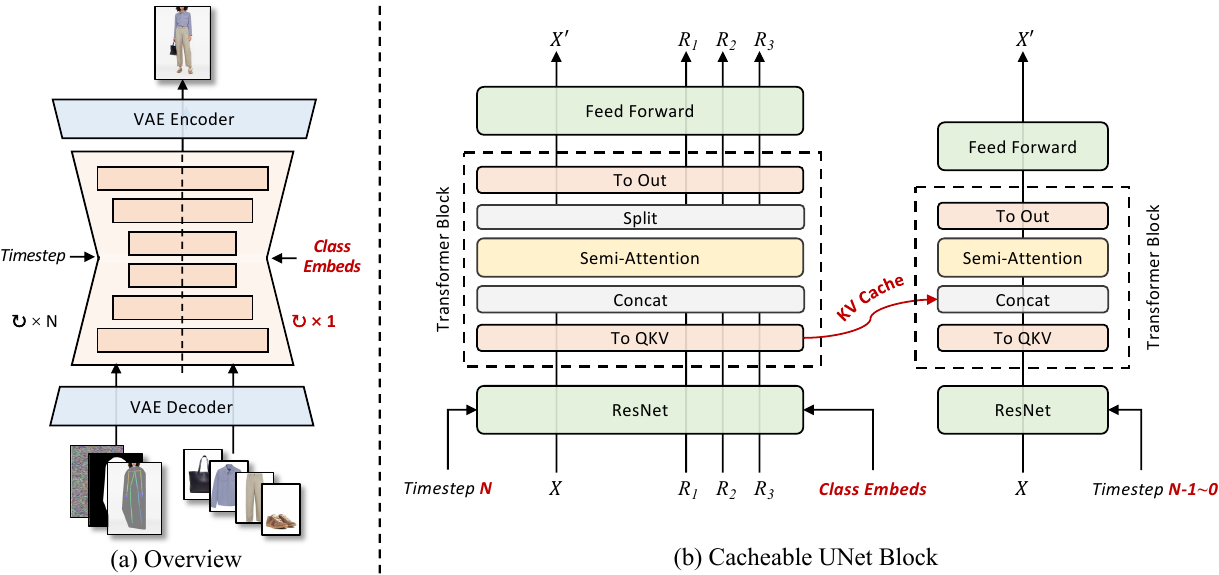}
  \caption{Overview of FastFit. Our model accelerates multi-reference virtual try-on through a novel cacheable UNet architecture. It enables lossless KV caching for reference features by conditioning them on class embeds instead of the timestep and using Semi-Attention to interact with the denoising features, which eliminates redundant computations.}
  \label{fig:framework}
\end{figure*}

\subsection{Overview}

The overall framework of FastFit is built upon the foundation of Latent Diffusion Models (LDMs) \citep{rombach2021ldm} and is designed to achieve high-speed, multi-reference virtual try-on through a novel conditioning cacheable UNet architecture. The entire workflow is depicted in \Cref{fig:framework} (a).
To ensure the generated image preserves the person's identity and pose while accurately rendering the new garments, we prepare two sets of conditions:

\vspace{1mm}
\noindent \textbf{Person Conditioning $c_p$}: To accurately preserve the person's identity and body pose, we construct the person condition $c_p$. First, we utilize AutoMask~\citep{chong2024catvtonconcatenationneedvirtual} to generate a cloth-agnostic mask $\mathbf{M_a}$ from the input image $I_p$. Subsequently, a composite image, $\mathbf{I_{comp}}$, is created by combining the human pose skeleton extracted via DWPose~\citep{yang2023dwpose} with the person image masked by $M_a$. $c_p$ is formed as: 
\begin{equation}
\label{eq:cp_condition}
c_p = \text{Concat}(\text{Interpolate}(M_a), \mathcal{E}(I_{comp}))
\end{equation}
where $\mathcal{E}$ is the VAE encoder, $\text{Interpolate}$ is a downsampling function that resizes the mask $M_a$, and $\text{Concat}$ denotes the channel-wise concatenation.

\vspace{1mm}
\noindent \textbf{Reference Conditioning $\{R_i\}_{i=1}^K$}: To capture the detailed appearance of the target garments, we extract a set of reference latents $\{R_i\}_{i=1}^K$ from the corresponding reference images $\{I_{R_i}\}_{i=1}^K$, which is defined as:
\begin{equation}
\label{eq:ref_condition}
\{R_i\}_{i=1}^K = \{\mathcal{E}(I_{R_i})\}_{i=1}^K
\end{equation}

\vspace{1mm}

The image generation process is guided by a denoising UNet $\epsilon_\theta$, which predicts the noise $\tilde{\epsilon}_t$ at each timestep $t$. As illustrated in \Cref{fig:framework} (a), our key innovation is to conceptually partition the function of $\epsilon_\theta$ into two streams: a time-independent path for reference inputs and a time-dependent path for the denoising process. 
Specifically, each reference latent $R_i$ is processed individually by a dedicated, time-independent path within the UNet, conditioned only on its corresponding Class Embedding $E_i$. This allows us to pre-compute and cache a separate feature representation, $\mathcal{R}_{\text{cache}}^{(i)}$, for each item before the denoising loop begins. This operation is performed for all $i \in \{1, \dots, K\}$ and is independent of any timestep $t$:
\begin{equation}
\label{eq:ref_cache}
\mathcal{R}_{\text{cache}}^{(i)} = \epsilon_\theta(R_i, E_i) \quad \text{for } i=1, \dots, K
\end{equation}
The resulting set of cached features, $\{\mathcal{R}_{\text{cache}}^{(i)}\}_{i=1}^K$, is then collectively used in each step of the main denoising loop.

The main denoising loop then proceeds for $N$ steps. At each step $t$, the UNet, $\epsilon_\theta$, processes only the time-dependent inputs: the noisy latent $z_t$, the person condition $c_p$, and the timestep embedding $\gamma(t)$. It integrates the static reference information by attending to the pre-computed set of cached features, $\{\mathcal{R}_{\text{cache}}^{(i)}\}_{i=1}^K$, via a Semi-Attention mechanism (detailed in \Cref{sec:cacheable_unet}):
\begin{equation}
\label{eq:denoise_step}
\tilde{\epsilon}_t = \epsilon_\theta(z_t, c_p, \gamma(t), \{\mathcal{R}_{\text{cache}}^{(i)}\}_{i=1}^K)
\end{equation}
This decomposition of the denoising process is the key to FastFit's efficiency, as it shifts the expensive computation for multiple reference features entirely out of the iterative loop. Once the process concludes at $t=0$, the final clean latent, $z_0$, is mapped back to the pixel space using the VAE decoder $\mathcal{D}$, to produce the high-resolution output image, $I_{\text{out}}$:
\begin{equation}
\label{eq:decode_step}
I_{\text{out}} = \mathcal{D}(z_0)
\end{equation}

\subsection{Cacheable UNet for Efficient Conditioning}
\label{sec:cacheable_unet}
The primary bottleneck in existing subject-driven diffusion models is the repeated computation of reference features at every denoising step. This is because the reference conditioning is typically dependent on the timestep $t$, making the features dynamic. Our key innovation, the Cacheable UNet, fundamentally breaks this dependency, enabling reference features to be computed once and reused. This is achieved through two core components: Reference Class Embedding and a Semi-Attention mechanism, as illustrated in \Cref{fig:framework} (b).

\vspace{-1mm}
\paragraph{Reference Class Embedding.}
To decouple the reference features from the denoising timestep $t$, we replace the conventional timestep embedding with a static, learnable Reference Class Embedding for the reference items. Specifically, for a set of $K$ reference items $\{R_1, \dots, R_K\}$, each belonging to a certain category (e.g., 'top', 'shoes'), we introduce a corresponding set of learnable class embeddings $\{E_1, \dots, E_K\}$. The features for each reference item $R_i$ are conditioned on its class embedding $E_i$ instead of the shared timestep embedding $\gamma(t)$ used by the denoising features $X$. The reference class embedding is injected in the same manner as the timestep embedding; both modulate the features within the ResNet blocks through an scaling operation. Since the class embeddings are constant throughout the entire denoising process, the resulting reference features become static and independent of the current timestep $t$, making them inherently cacheable. 

\begin{wrapfigure}{tr}{0.39\textwidth}
  \vspace{-10pt} 
  \centering
  \includegraphics[width=0.84\linewidth]{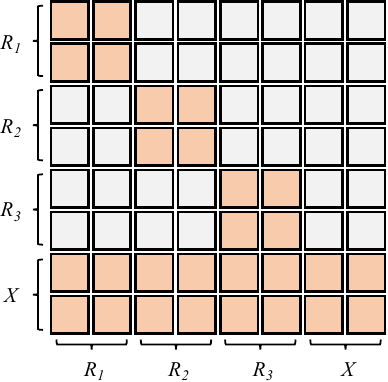}
  \vspace{-1mm}
 \caption{Visualization of the Semi-Attention Mask. Denosing feature $X$ attend to all features, while each reference feature $R_i$ is restricted to its own.}
\label{fig:semi_attention_mask}
\end{wrapfigure}

\vspace{-1mm}
\paragraph{Semi-Attention Mechanism.}
Having made reference features static, we need a mechanism to inject their information into the denoising process without compromising their static nature. A standard full self-attention would allow information to flow from the step-dependent denoising features $X$ back to the reference features $R_i$, thereby "contaminating" them and breaking the condition for caching.
To solve this, we propose a Semi-Attention mechanism, visualized in \Cref{fig:semi_attention_mask}. In this design, we treat both the denoising features $X$ and all reference features $\{R_i\}$ as a single sequence of tokens. The attention calculation is governed by a specific mask that controls the information flow: 
\textbf{(1) Denoising-to-All:} The tokens of the denoising feature $X$ are allowed to attend to all tokens in the sequence (i.e., to itself and to all reference features $R_i$). This allows the model to effectively "read" the appearance information from each garment and apply it to the person. 
\textbf{(2) Reference-to-Self:} The tokens of each reference feature $R_i$ are only allowed to attend to themselves. They cannot attend to the denoising features $X$ or to any other reference feature $R_j$ (where $j \neq i$).
This attention mask ensures that the reference features act as a static, read-only source of information for the denoising process. Their representations are never updated by the dynamic features of $X$, thus preserving their cacheability across all timesteps.

\vspace{1.5mm}
In summary, the Reference Class Embedding makes the computation of reference features static, while the Semi-Attention mechanism ensures that during interaction, the static reference features only provide information without being affected by the denoising process. This synergistic design forms the Cacheable UNet architecture, laying the foundation for an efficient, cache-based inference pipeline.

\subsection{Inference Acceleration with Reference KV Cache}
The design of our Cacheable UNet enables a highly efficient inference pipeline via a Reference KV Cache. As depicted in \Cref{fig:framework}(b), the process is split into two stages:

\paragraph{Pre-computation and Caching (One-time Cost).}
Before the iterative denoising loop begins, we perform a single forward pass for each reference item $R_i$ through the UNet $\epsilon_\theta$. For each Semi-Attention layer, we then compute and store its corresponding Key ($K_i^{\text{cache}}$) and Value ($V_i^{\text{cache}}$) matrices. This pre-computation step is performed only once per generation request.

\paragraph{Accelerated Denoising Loop.}
For every subsequent denoising step from $t=N-1$ down to $0$, we completely bypass the computation for the reference branches. Instead, for each Semi-Attention layer, we only compute the Query ($Q_X$), Key ($K_X$), and Value ($V_X$) from the current denoising features $X_t$. We then construct the full key and value matrices, $K_{\text{full}}$ and $V_{\text{full}}$, by concatenating these dynamic tensors with all the cached keys $\{K_i^{\text{cache}}\}_{i=1}^K$ and values $\{V_i^{\text{cache}}\}_{i=1}^K$, respectively:
\begin{align}
\label{eq:k_full_concat}
K_{\text{full}} &= \text{Concat}(K_X, K_1^{\text{cache}}, \dots, K_K^{\text{cache}}) \\
\label{eq:v_full_concat}
V_{\text{full}} &= \text{Concat}(V_X, V_1^{\text{cache}}, \dots, V_K^{\text{cache}})
\end{align}
The final attention output is then calculated only for the denoising query $Q_X$:
\begin{equation}
\label{eq:attention_with_cache}
\text{Attention}(Q_X, K_{\text{full}}, V_{\text{full}}) = \text{softmax}\left(\frac{Q_X K_{\text{full}}^T}{\sqrt{d_k}}\right) V_{\text{full}}
\end{equation}

\vspace{1.5mm}
This strategy effectively reduces the computational cost of attention at each step to be dependent only on the denoising features, regardless of the number or complexity of reference items. This fundamentally resolves the efficiency bottleneck, leading to a substantial reduction in inference latency, especially in the multi-reference setting central to our work.

\begin{figure*}
  \centering
  \includegraphics[width=\textwidth]{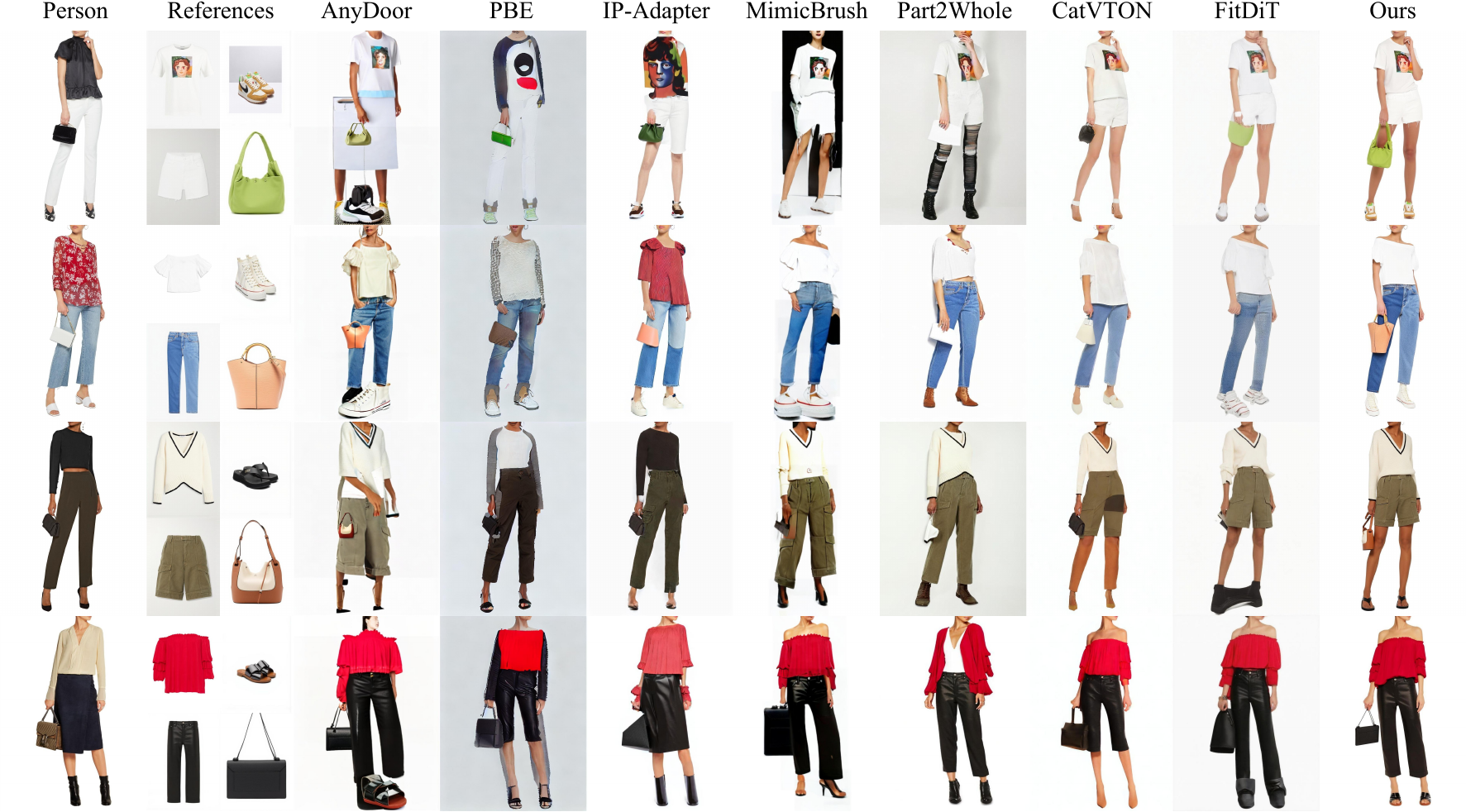}
    \vspace{-5mm}
  \caption{Qualitative comparison on the DressCode-MR dataset. See \Cref{sec:appendix_mr} for more results. Please zoom in for more details.}
  \label{fig:comparison_mr}
\end{figure*}

\section{Experiments}

\subsection{Datasets}
\label{sec:datasets}
We evaluate our model on three datasets, VITON-HD~\citep{choi2021vitonhd}, DressCode~\citep{morelli2022dresscode}, and our newly proposed DressCode-MR, all at 1024×768 resolution. VITON-HD~\citep{choi2021vitonhd} provides 13,679 image pairs for upper-body virtual try-on (11,647 train / 2,032 test). DressCode~\citep{morelli2022dresscode} dataset features 53,792 full-body pairs (48,392 train / 5,400 test) covering tops, bottoms, and dresses. 
To facilitate multi-reference research, we introduce DressCode-MR, built upon DressCode. As illustrated in \Cref{fig:dataset}, it contains 28,179 samples (25,779 train / 2,400 test), each pairing a person with a complete outfit from up to five categories: tops, bottoms, dresses, shoes, and bags.
We constructed this dataset by training five expert restoration models (based on CatVTON~\citep{chong2024catvtonconcatenationneedvirtual} and FLUX~\citep{flux2024}) using VITON-HD, DressCode, and a small set of internet-sourced shoe and bag pairs. These models were used to recover the canonical images for items worn in DressCode, and the final high-quality samples were selected through human feedback.


\begin{figure*}
  \centering
  \includegraphics[width=\textwidth]{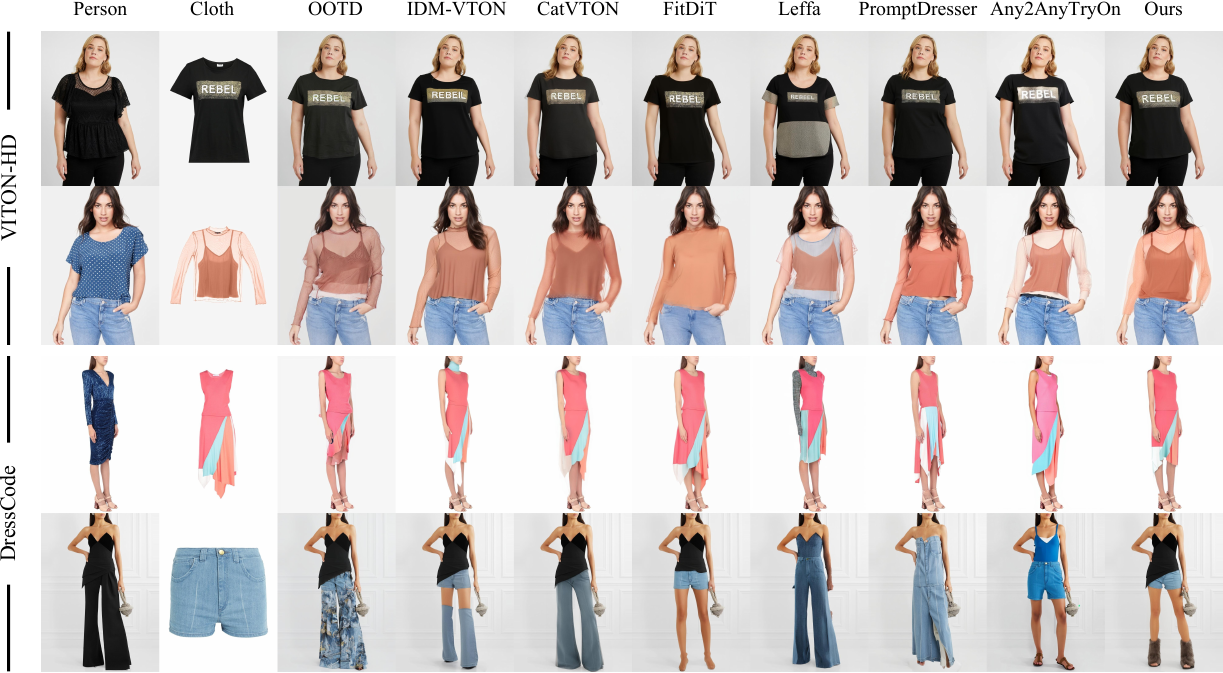}
    \vspace{-5mm}
\caption{Qualitative comparison on the VITON-HD \citep{choi2021vitonhd} and DressCode \citep{morelli2022dresscode} dataset. See \Cref{sec:appendix_sr} for more results. Please zoom in for more details.}
    \label{fig:comparison_sr}
\end{figure*}

\subsection{Implementation Details}
\label{imple_details}
We train our single-reference try-on model based on the pretrained StableDiffusion v1.5~\citep{rombach2021ldm} inpainting on the DressCode~\citep{morelli2022dresscode} and VITON-HD~\citep{choi2021vitonhd} datasets for 64,000 steps with a batch size of 32 and a resolution of 1024$\times$768. This version is used for all single-reference quantitative evaluations. Building upon the single-reference model, we fine-tune it on our proposed DressCode-MR dataset for 16,000 steps with the same resolution and batch size. We utilized the AdamW~\citep{loshchilov2019adamw} optimizer with a constant learning rate of $1\times10^{-5}$ for both training stages. To enable classifier-free guidance, 20\% of the reference images were randomly dropped during the training. All experiments were conducted on 8 NVIDIA H100 GPUs.

\subsection{Metrics}
We evaluate our model's performance on two fronts: image fidelity and computational efficiency. 

\vspace{1mm}
\noindent \textbf{Image Fidelity.} We use two settings. In the paired setting, where ground-truth images are available, we measure similarity using the Structural Similarity Index (SSIM)~\citep{wang2004ssim}, Learned Perceptual Image Patch Similarity (LPIPS)~\citep{zhang2018LPIPS}, Fréchet Inception Distance (FID)~\citep{Seitzer2020FID}, and Kernel Inception Distance (KID)~\citep{bińkowski2021kid}. In the unpaired setting, we assess overall realism and diversity by comparing the distribution of our generated samples to that of real images using FID and KID. 

\vspace{1mm}
\noindent \textbf{Computational Efficiency.} We report the total parameters, inference latency, and peak memory usage. These metrics are benchmarked by averaging 100 runs on a single NVIDIA H100 GPU, with each run configured for 20 denoising steps and with classifier-free guidance (CFG)~\citep{ho2022classifierfreediffusionguidance} enabled.

\newpage

\subsection{Quantitative Comparison}
\label{sec:quantity}

\begin{wraptable}{rt}{0.45\textwidth}
    \centering
    \small
    \vspace{-4.8mm}
    \caption{Quantitative comparison of model efficiency. Best and second-best results are in \textbf{bold} and \underline{underlined}, respectively.}
    \resizebox{\linewidth}{!}{
    \setlength{\tabcolsep}{1mm}
    \begin{tabular}{lccc}
    \toprule
    Method          & Params(M)$\downarrow$ & Time(s)$\downarrow$ & Memory(M)$\downarrow$ \\
    \midrule
    Any2AnyTryon    & 16786.78   & 12.19    & 35218      \\
    PromptDresser   &  6011.03   &  4.29    & 17364      \\
    FitDiT          &  5870.80   &  2.00    & 15992      \\
    Leffa           &  1802.72   &  3.32    &  7996      \\
    IDM-VTON        &  7086.91   &  2.76    & 19072      \\
    OOTDiffusion    &  2229.73   &  \underline{1.93}    & 10154      \\
    CatVTON         &   \textbf{899.06}   &  2.10    &  \textbf{5500}      \\
    \midrule
    \rowcolor{LightBlue} \textbf{FastFit}         &   \underline{904.86}   &  \textbf{1.16}    &  \underline{6944}      \\
    \bottomrule
    \end{tabular}
    }
    \vspace{-2mm}
    \label{tab:efficient}
\end{wraptable}

\paragraph{Single-Reference Virtual Try-On.}
We conducted a quantitative comparison against current state-of-the-art virtual try-on methods \citep{guo2025any2anytryonleveragingadaptiveposition, kim2024promptdresserimprovingqualitycontrollability, jiang2024fitditadvancingauthenticgarment, choi2024idmvton, chong2024catvtonconcatenationneedvirtual, xu2024ootdiffusion} on VITON-HD \citep{choi2021vitonhd} and DressCode \citep{morelli2022dresscode} datasets. 
As shown in \Cref{tab:combined_quantitative}, FastFit achieves competitive results across both datasets under paired and unpaired settings, demonstrating its superior capability in generating high-quality images.
\Cref{tab:efficient} highlights the efficiency of FastFit, which achieves an average 3.5$\times$ speedup over comparable methods while remaining competitive in terms of parameters and memory usage.

\begin{wraptable}{rt}{0.61\textwidth}
    \centering
    \small
    \setlength{\tabcolsep}{0.7mm} 

    \vspace{-4.5mm}
    \caption{Quantitative comparison on DressCode-MR for multi-reference try-on. Best and second-best results are in \textbf{bold} and \underline{underlined}, respectively.}
    \resizebox{\linewidth}{!}{
    \begin{tabular}{l c cccc cc}
        \toprule
        \multirow{2}{*}{Method} & \multirow{2}{*}{Time(s)$\downarrow$} & \multicolumn{4}{c}{Paired} & \multicolumn{2}{c}{unpair} \\
        \cmidrule(lr){3-6} \cmidrule(lr){7-8}
        & & FID$\downarrow$ & KID$\downarrow$ & SSIM$\uparrow$ & LPIPS$\downarrow$ & FID$\downarrow$ & KID$\downarrow$ \\
        \midrule
        AnyDoor          & 12.08 & 37.138 & 22.571 & 0.768 & 0.235 & 44.068 & 23.958 \\
        Paint-By-Example & 5.22  & 28.296 & 16.092 & 0.796 & 0.215 & 31.135 & 17.887 \\
        MimicBrush       & 6.62  & 21.074 & 9.858  & 0.800 & 0.173 & 22.111 & 9.992 \\
        Part2Whole       & 5.73  & 20.313 & 8.200  & 0.807 & 0.187 & 24.564 & 10.581 \\
        CatVTON          & 8.94  & 16.131 & 6.980  & 0.856 & 0.106 & 18.339 & 7.458 \\
        IP-Adapter       & 5.62  & \underline{14.459} & \underline{4.144} & \textbf{0.861} & \underline{0.089} & 24.139 & 10.783 \\
        FitDIT           & \underline{3.38}  & 14.722 & 5.471 & 0.850 & 0.122 & \underline{15.956} & \underline{5.645} \\
        \midrule
        \rowcolor{LightBlue}
        \textbf{FastFit} & \textbf{1.90} & \textbf{9.311} & \textbf{1.512} & \underline{0.859} & \textbf{0.079} & \textbf{12.059} & \textbf{2.123} \\
        \bottomrule
    \end{tabular}
    }
    \label{tab:multi_ref}
    \vspace{-2mm}
\end{wraptable}

\paragraph{Multi-Reference Virtual Try-On.}
\Cref{tab:multi_ref} shows our multi-reference try-on results. In the absence of methods designed for simultaneous multi-reference generation, we adapt strong baselines from subject-driven generation \citep{ye2023ip-adapter, yang2022paintbyexample, chen2023anydoor, chen2024mimicbrush} and multi-category try-on \citep{jiang2024fitditadvancingauthenticgarment, chong2024catvtonconcatenationneedvirtual, huang2024parts2whole} via sequential single-reference inference. FastFit achieves state-of-the-art scores across quality metrics and is also the most efficient method. This demonstrates its superior ability to cohesively synthesize multiple references with high fidelity.

\subsection{Qualitative Comparison}

\vspace{1mm}
\noindent \textbf{Single-Reference Virtual Try-On.}
\Cref{fig:comparison_sr} shows the qualitative comparison for the single-reference try-on task. On the VITON-HD \citep{choi2021vitonhd} dataset, our method excels at preserving fine-grained details, such as the text ``REBEL'' on T-shirts, where other methods often produce blurred results. FastFit also realistically renders challenging materials, like the sheer polka-dot top. On the DressCode \citep{morelli2022dresscode} dataset, our model accurately captures the correct shape and style of complex garments like the high-slit dress.

\vspace{1mm}
\noindent \textbf{Multi-Reference Virtual Try-On.}
We further evaluate FastFit on the more challenging multi-reference virtual try-on task, with results presented in \Cref{fig:comparison_mr}. The comparison clearly demonstrates our model's superior capability. FastFit successfully synthesizes a coherent and realistic final image by seamlessly combining multiple reference items. In contrast, most existing methods, such as AnyDoor \citep{chen2023anydoor} and PBE \citep{yang2022paintbyexample}, often fail to properly compose the different garments or produce significant artifacts. Our method, however, maintains the identity and details of each piece of clothing, resulting in a natural and believable complete outfit.

\begin{table*}[t]
    \centering
    \small
    \caption{Quantitative comparison for single-reference virtual try-on on the VITON-HD~\citep{choi2021vitonhd} and DressCode~\citep{morelli2022dresscode} datasets. All metrics are rounded to three decimal places. Best and second-best results in each column are in \textbf{bold} and \underline{underlined}, respectively.}
    \label{tab:combined_quantitative}
    \resizebox{\textwidth}{!}{
    \setlength{\tabcolsep}{1.5mm} 
    \begin{tabular}{lcccccccccccc}
        \toprule
        \multirow{2}{*}{Method} & \multicolumn{6}{c}{VITON-HD} & \multicolumn{6}{c}{DressCode} \\
        \cmidrule(lr){2-7} \cmidrule(lr){8-13}
        & \multicolumn{4}{c}{Paired} & \multicolumn{2}{c}{Unpaired} & \multicolumn{4}{c}{Paired} & \multicolumn{2}{c}{Unpaired} \\
        \cmidrule(lr){2-5} \cmidrule(lr){6-7} \cmidrule(lr){8-11} \cmidrule(lr){12-13}
        & FID $\downarrow$ & KID $\downarrow$ & SSIM $\uparrow$ & LPIPS $\downarrow$ & FID $\downarrow$ & KID $\downarrow$ & FID $\downarrow$ & KID $\downarrow$ & SSIM $\uparrow$ & LPIPS $\downarrow$ & FID $\downarrow$ & KID $\downarrow$ \\
        \midrule
        Any2AnyTryon & 11.195 & 2.806 & 0.799 & 0.194 & 9.981 & 3.496 & 5.111 & 1.265 & 0.897 & 0.059 & 6.709 & 1.580 \\
        PromptDresser & 5.934 & 0.550 & 0.846 & 0.090 & 8.885 & 0.909 & 9.563 & 4.795 & 0.858 & 0.104 & 10.618 & 4.978 \\
        FitDiT & 8.176 & 1.079 & 0.838 & 0.096 & 9.979 & 1.478 & 5.571 & 1.901 & 0.899 & \underline{0.058} & \underline{4.805} & \underline{0.712} \\
        Leffa & \underline{5.667} & 0.692 & 0.857 & 0.076 & 10.446 & 2.640 & 7.193 & 2.114 & 0.861 & 0.084 & 20.099 & 13.506 \\
        IDM-VTON & 6.112 & 1.112 & 0.866 & \underline{0.074} & 9.249 & 1.267 & 7.181 & 3.524 & 0.891 & 0.070 & 9.167 & 4.489 \\
        CatVTON & 6.738 & 1.320 & \underline{0.881} & 0.088 & 10.552 & 2.272 & \underline{3.710} & \underline{1.010} & \textbf{0.909} & 0.062 & 5.872 & 1.606 \\
        OOTDiffusion & 5.762 & \textbf{0.267} & 0.843 & \textbf{0.072} & \underline{9.082} & \underline{0.702} & 6.975 & 2.014 & 0.873 & 0.077 & 8.121 & 2.886 \\
        \midrule
        \rowcolor{LightBlue} \textbf{FastFit} & \textbf{5.629} & \underline{0.505} & \textbf{0.885} & 0.078 & \textbf{8.629} & \textbf{0.665} & \textbf{2.836} & \textbf{0.390} & \underline{0.907} & \textbf{0.057} & \textbf{4.397} & \textbf{0.553} \\
        \bottomrule
    \end{tabular}
    }
\end{table*}

\subsection{Ablation Studies}
\label{sec:ablation}

\begin{table*}[t]
    \centering
    \small
    \setlength{\tabcolsep}{2.2mm}
    \caption{Ablation study of the key components in our model on DressCode~\citep{morelli2022dresscode} dataset. The best and second-best results are demonstrated in \textbf{bold} and \underline{underlined}, respectively.
    }
    \resizebox{\textwidth}{!}{
        \begin{tabular}{lccccccccc}
        \toprule
        \multirow{2}{*}{Variants} & \multirow{2}{*}{Params(M)$\downarrow$} & \multirow{2}{*}{Time (s)$\downarrow$} & \multirow{2}{*}{Memory (M)$\downarrow$} & \multicolumn{4}{c}{Paired} & \multicolumn{2}{c}{Unpaired} \\
        \cmidrule(lr){5-8} \cmidrule(lr){9-10}
        & & & & FID$\downarrow$ & KID$\downarrow$ & SSIM$\uparrow$ & LPIPS$\downarrow$ & FID$\downarrow$ & KID$\downarrow$ \\
        \midrule
        w/o KV Cache        & \underline{904.86}       & \underline{1.92} & \textbf{6944}   &  \underline{2.8585}  & \underline{0.3737} & \textbf{0.9057} & \textbf{0.0588} & \textbf{4.4206} & \underline{0.5903} \\
        w/ Full Attention   & \underline{904.86}       & 2.17 & \textbf{6944}    & 3.1847  & 0.5426 & \underline{0.9056} & 0.0606 & 4.6221       & 0.6533      \\
        w/o Class Embed     & \textbf{904.85}  & \textbf{1.16} & \textbf{6944}    & 2.9146  & 0.4000 & \underline{0.9056} & \underline{0.0591} & 4.4624 & 0.5929 \\
        w/ ReferenceNet     & 1729.92 & \textbf{1.16}    & \underline{8770} & \textbf{2.8474}  & \textbf{0.3577} & 0.9054 & \textbf{0.0588} & \underline{4.4365} & \textbf{0.5741} \\
        \midrule
        \rowcolor{LightBlue}
        \textbf{FastFit}    & \underline{904.86}  & \textbf{1.16} & \textbf{6944} & \underline{2.8585}  & \underline{0.3737} & \textbf{0.9057} & \textbf{0.0588} & \textbf{4.4206} & \underline{0.5903} \\
        \bottomrule
        \end{tabular}
    }
    \vspace{-1mm}
    \label{tab:fastfit_ablation}
\end{table*}

The results in \Cref{tab:fastfit_ablation} validate our key design choices. Firstly, the Reference KV Cache is crucial for efficiency; disabling it increases inference time from 1.16s to 1.92s, yet this $\sim$1.66$\times$ speedup comes with no loss in generation quality, as the performance metrics are identical. Secondly, our parameter-sharing strategy is highly effective. Introducing a separate ReferenceNet nearly doubles the parameters (904.86M $\rightarrow$ 1729.9M) and increases memory usage, but yields no corresponding performance improvement. Furthermore, replacing Semi-Attention with Full Attention is detrimental, as it not only slows inference to 2.17s but also degrades generation quality (e.g., FID increases to 3.1847). We hypothesize this is because full interaction disrupts the consistency of reference features. Finally, removing the Class Embedding causes a slight performance drop, and its effectiveness in guiding region-specific attention is presented in \Cref{sec:appendix_ablation_class_embed}. All ablation experiments follow the settings described in \Cref{imple_details}, trained for 32K steps, and are evaluated on the DressCode~\citep{morelli2022dresscode} dataset.

\section{Conclusion}
In this paper, we proposed FastFit, a high-speed multi-reference virtual try-on framework designed to break the critical trade-offs between versatility, efficiency, and quality in existing technologies. Through an innovative Cacheable UNet, which combines a Class Embedding and a Semi-Attention mechanism, we decoupled reference feature encoding from the denoising process. This design enables a Reference KV Cache that allows reference features to be computed once and reused losslessly across all steps, fundamentally eliminating the computational redundancy that plagues current methods. Experimental results show that FastFit achieves a significant efficiency advantage—an average 3.5$\times$ speedup over comparable methods—without sacrificing generation quality. For the first time, it enables coherent, synergistic try-on for up to 5 key categories: tops, bottoms, dresses, shoes, and bags. Furthermore, the DressCode-MR dataset we constructed provides a valuable foundation for future research in complex outfit generation. In summary, FastFit represents a promising advance towards a more realistic, efficient, and diverse virtual try-on experience, significantly lowering the barriers for its widespread application in e-commerce and intelligent outfit visualization.

\paragraph{Limitations and Future Work.}
Despite the model's strong performance, several areas present opportunities for future exploration. To further enhance realism, the modeling of complex physical interactions and layering among garments could be improved. Expanding the DressCode-MR dataset with such complex interaction pairs would be a valuable direction. Another important research path is improving generalization to underrepresented apparel, such as styles with unique topologies or challenging materials. Finally, while our framework significantly accelerates inference, a gap remains toward achieving real-time interaction. Exploring techniques such as guidance and step distillation, combined with more advanced caching mechanisms, offers a promising path to bridge this gap and enable applications like interactive real-time outfit visualization.


\bibliography{iclr2025_conference}
\bibliographystyle{iclr2025_conference}

\newpage
\appendix

\section{Appendix}

\subsection{Quantitative Comparison across Garment Types}

For a more fine-grained analysis, \Cref{tab:quant_compare_dresscode} presents a quantitative comparison on the DressCode \citep{morelli2022dresscode} dataset, with results broken down by clothing category. The results highlight the robust and superior performance of our method across all tested categories, including upper, lower, and dresses. FastFit consistently achieves either the best or second-best scores in the vast majority of key metrics, demonstrating its strong and stable performance regardless of the garment type. This showcases the model's excellent generalization capability for different clothing styles.

\begin{table*}[h]
    \centering
    \small
    \setlength{\tabcolsep}{1.2mm} 
    \caption{Quantitative comparison on the DressCode dataset, with results broken down by category (Upper, Lower, and Dresses). The best results are marked in \textbf{bold} and the second-best are \underline{underlined}. $\downarrow$ indicates lower is better, while $\uparrow$ indicates higher is better.}
    \label{tab:quant_compare_dresscode}
    \resizebox{\textwidth}{!}{ 
        \begin{tabular}{l cccc cccc cccc}
        \toprule
        \multirow{2}{*}{Methods} & \multicolumn{4}{c}{Upper} & \multicolumn{4}{c}{Lower} & \multicolumn{4}{c}{Dresses} \\
        \cmidrule(lr){2-5} \cmidrule(lr){6-9} \cmidrule(lr){10-13}
        & FID$\downarrow$ & KID$\downarrow$ & SSIM$\uparrow$ & LPIPS$\downarrow$ & FID$\downarrow$ & KID$\downarrow$ & SSIM$\uparrow$ & LPIPS$\downarrow$ & FID$\downarrow$ & KID$\downarrow$ & SSIM$\uparrow$ & LPIPS$\downarrow$ \\
        \midrule
        Any2AnyTryon & 10.4741 & 1.7130 & 0.9206 & 0.0476 & 13.1152 & 2.8336 & 0.8896 & 0.0655 & 9.1124 & 1.6539 & 0.8796 & 0.0636 \\
        PromptDresser & 9.2447 & 0.7174 & 0.9044 & 0.0678 & 32.9093 & 17.9749 & 0.8327 & 0.1352 & 16.8179 & 6.8932 & 0.8363 & 0.1087 \\
        Leffa & 11.2549 & 2.0947 & 0.8908 & 0.0578 & 19.6834 & 5.8985 & 0.8594 & 0.0908 & 13.3859 & 2.3701 & 0.8335 & 0.1029 \\
        IDM-VTON & 11.2283 & 3.3860 & 0.9174 & 0.0547 & 11.7878 & 3.4218 & 0.8978 & 0.0655 & 17.5135 & 8.3389 & 0.8585 & 0.0882 \\
        OOTDiffusion & 9.5945 & 1.1055 & 0.9040 & 0.0528 & 19.6615 & 6.1217 & 0.8751 & 0.0827 & 14.8496 & 4.4567 & 0.8393 & 0.0963 \\
        CatVTON & \underline{7.8465} & 1.0851 & \textbf{0.9360} & 0.0504 & \underline{8.6135} & \underline{1.6574} & \textbf{0.9236} & \underline{0.0562} & 8.9453 & 1.0575 & 0.8669 & 0.0791 \\
        FitDiT & 8.0876 & \underline{0.5789} & 0.9241 & \textbf{0.0417} & 24.5079 & 11.7225 & 0.8944 & 0.0758 & \textbf{7.2253} & \underline{0.4768} & \textbf{0.8789} & \textbf{0.0562} \\
        \midrule
        \rowcolor{LightBlue}
        \textbf{FastFit} & \textbf{6.8354} & \textbf{0.2453} & \underline{0.9318} & \underline{0.0485} & \textbf{7.1311} & \textbf{0.7981} & \underline{0.9207} & \textbf{0.0511} & \underline{7.5890} & \textbf{0.2446} & \underline{0.8671} & \underline{0.0720} \\
        \bottomrule
        \end{tabular}
    }
    \vspace{-1mm} 
\end{table*}

\subsection{More Visual Comparisons}

\subsubsection{Single-Reference Virtual Try-On}
\label{sec:appendix_sr}
\begin{figure*}
  \centering
  \includegraphics[width=1.0\textwidth]{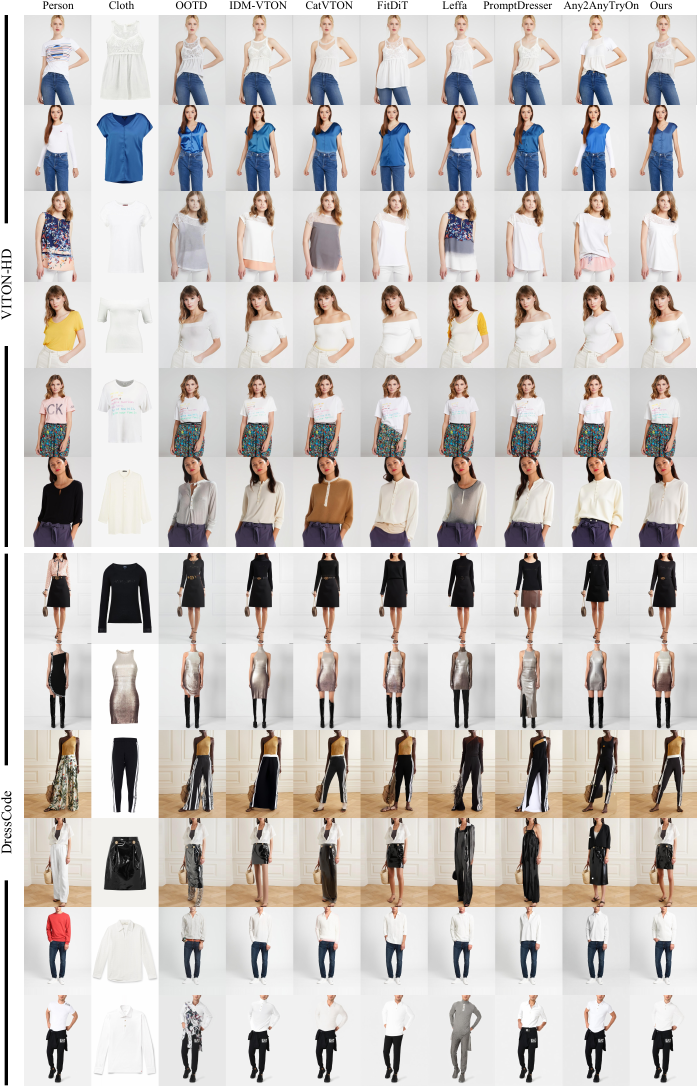}
  \caption{More visual comparisons on the VITON-HD \citep{choi2021vitonhd} and DressCode \citep{morelli2022dresscode} dataset with baseline methods. Please zoom in for more details.}
  \label{fig:appendix_sr}
\end{figure*}

In the single-reference virtual try-on task, our method demonstrates robust performance across both the VITON-HD \citep{choi2021vitonhd} and DressCode \citep{morelli2022dresscode} datasets. As illustrated in \Cref{fig:appendix_sr}, FastFit excels at preserving high-frequency details on the garments, such as intricate patterns and text logos. Furthermore, our model accurately renders the correct shape and length for various types of clothing. The final results show that the garments are naturally fused with the person's body, effectively handling challenging poses and occlusions.

\subsubsection{Multi-Reference Virtual Try-On} 
\label{sec:appendix_mr}

\begin{figure*}
  \centering
  \includegraphics[width=1.0\textwidth]{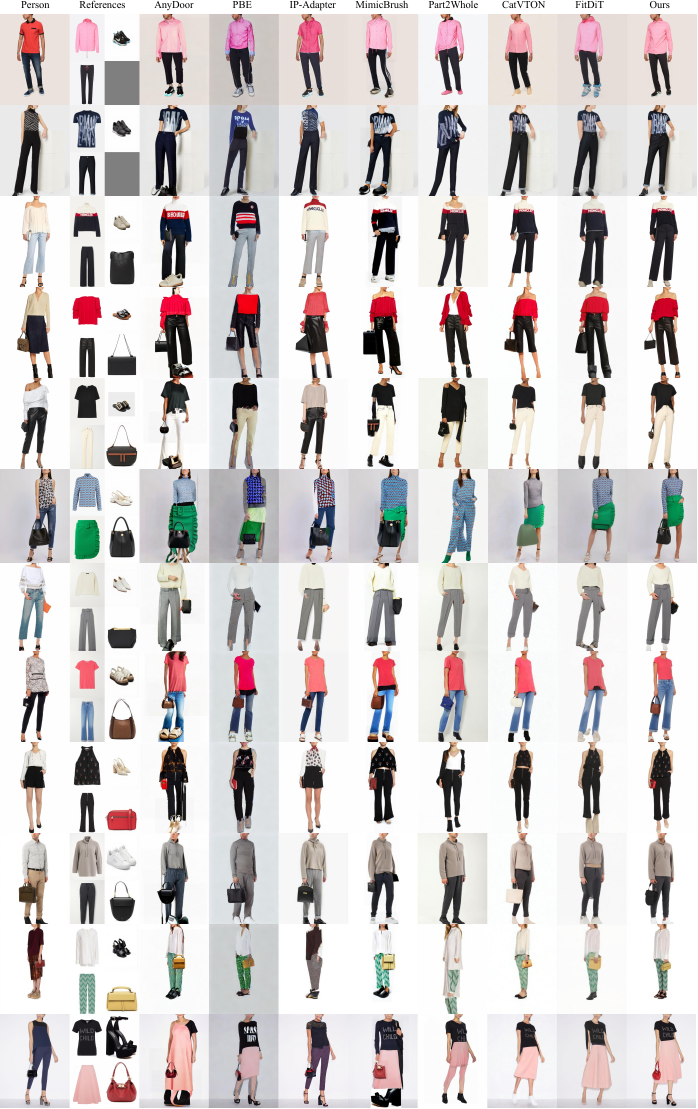}
  \caption{More visual comparisons on the DressCode-MR dataset with baseline methods. Please zoom in for more details.}
  \label{fig:appendix_mr}
\end{figure*}

For the more challenging multi-reference task, FastFit exhibits a significant advantage over competing methods. \Cref{fig:appendix_mr} showcases our model's unique ability to seamlessly combine multiple, distinct reference items into a single, coherent outfit. Notably, even during this complex composition process, FastFit faithfully preserves the fine-grained details and logos of each individual item (e.g., "SPAS", "CHIUS"). This capability to generate complete and detailed ensembles in complex scenarios highlights its superiority where other methods often struggle.

\subsection{Visual Analysis of the Effectiveness of Class Embeddings}
\label{sec:appendix_ablation_class_embed}

\begin{figure*}
  \centering
  \includegraphics[width=\textwidth]{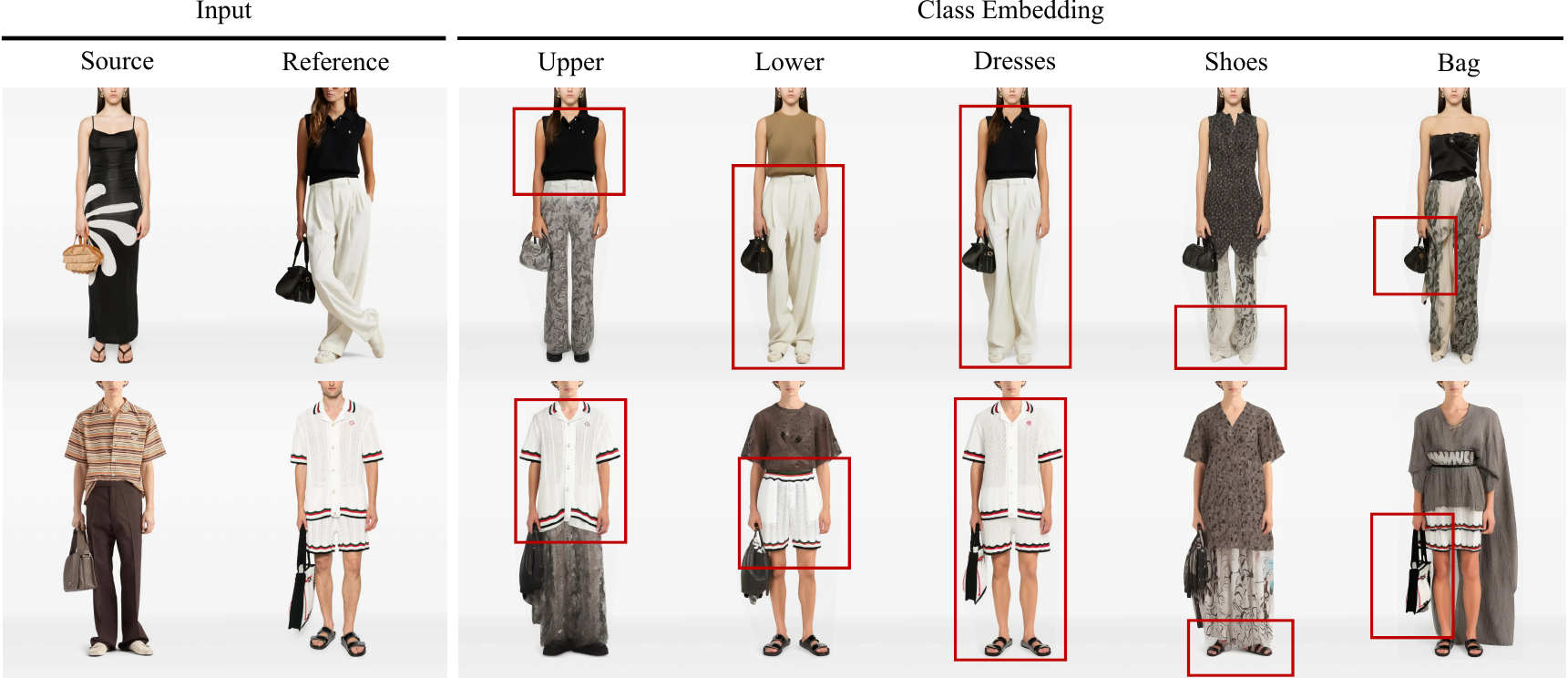}
  \caption{Demonstration of the visual effect of class embeddings. By providing different class embeddings (e.g., 'Upper', 'Lower', 'Dresses', 'Shoes', 'Bag') for the same reference image, our model can be directed to selectively transfer the corresponding item to the source person.}
  \label{fig:ablation_class_embed}
\end{figure*}

To visually validate the effectiveness of the Reference Class Embedding as a key control mechanism in our model, we conducted an additional ablation study. As shown in \Cref{fig:ablation_class_embed}, the experiment is designed to isolate the influence of the class embedding. For each example row, we provide the model with the exact same source person and reference image. The only variable changed across the columns is the specific class embedding provided (e.g., 'Upper', 'Lower', 'Dresses', 'Shoes', 'Bag').
The results demonstrate that the class embedding provides fine-grained, semantic control over the try-on process. The model is able to interpret the embedding and selectively transfer the corresponding item from the reference image, even when multiple items are present. 
This experiment confirms that by applying a Class Embedding, the model's attention is effectively guided to the corresponding region of the reference image, which is crucial for preventing the features of different reference items from being conflated in a multi-reference scenario.

\end{document}